\documentclass[10 pt, conference]{IEEEtran}

\usepackage{cite}
\usepackage{amsmath,amssymb,amsfonts}
\usepackage{algorithmic}
\usepackage{graphicx}
\usepackage{textcomp}
\usepackage{xcolor}
\usepackage{lipsum}
\usepackage{hyperref}
\def\BibTeX{{\rm B\kern-.05em{\sc i\kern-.025em b}\kern-.08em
    T\kern-.1667em\lower.7ex\hbox{E}\kern-.125emX}}
\IEEEoverridecommandlockouts
\begin{document}

\title{Automatic Eye-in-Hand Calibration using EKF\\
\vspace{0.2in}
\author{
  \IEEEauthorblockN{%
    Aditya Ramakrishnan\IEEEauthorrefmark{1},
    Chinmay Garg\IEEEauthorrefmark{1},
    Haoyang He\IEEEauthorrefmark{1},
    Shravan Kumar Gulvadi\IEEEauthorrefmark{1} and
    Sandeep Keshavegowda%
  }%
  
\vspace{0.1in}
\IEEEauthorblockA{Carnegie Mellon University, Pittsburgh, PA, USA}

\vspace{0.1in}
\IEEEauthorblockA{Link to project code: \href{https://github.com/nefario7/hec-slam}{hec-slam}}

\thanks{\IEEEauthorrefmark{1} Authors contributed equally}
\thanks{Carnegie Mellon University, Pittsburgh, PA, USA
        {\tt \{aramkr2, chinmayg, hhe2, sgulvadi, skeshave\}@andrew.cmu.edu}}%
}

}

\maketitle

\begin{abstract}
In this paper, a self-calibration approach for eye-in-hand robots using SLAM is considered. The goal is to calibrate the positioning of a robotic arm, with a camera mounted on the end-effector automatically using a SLAM-based method like Extended Kalman Filter (EKF). Given the camera intrinsic parameters and a set of feature markers in a work-space, the camera extrinsic parameters are approximated. An EKF based measurement model is deployed to effectively localize the camera and compute the camera to end-effector transformation. The proposed approach is tested on a UR5 manipulator with a depth-camera mounted on the end-effector to validate our results.
\end{abstract}

\section{Introduction}
Autonomous robotic platforms today are becoming increasingly ubiquitous in a wide array of domains. An integral feature of modern autonomous systems is the robot's ability to interpret external information through sophisticated sensors that are susceptible to environmental noise and disruptions that impact the sensor extrinsic properties.\\

It is imperative to understand that for true long-term autonomy applications, robust self-calibration is essential as changes in sensor extrinsics have a significant impact on the ability of the robot to localize accurately. Conventional processes involve sophisticated calibration procedures, prior to sensor-data processing or even robot deployment, in order to estimate extrinsic parameters to a high degree of accuracy, a feature point of these methods being the lack of sensor calibration updation, till end of use.\\

The motivation behind this project is to remove the tedious calibration procedure from robotic applications and to enable online auto-calibration that adapts to future industry standards. Optimizing efficient performance of a robotic arm in common Pick and Place applications requires the robot to know the relative pose of the camera with respect to itself and thereby map the position and orientation of the objects from the image to the real world. This calibration process is aptly called ‘Eye-in-Hand Calibration’ (since the camera is mounted at the end-effector of the robotic arm).\\

Conventional methods of camera calibration requires the prior knowledge of key-point positions in the global frame of reference which is then used to calculate the camera position. This approach is tedious and can prove to be severely restrictive as it requires a template with points mapped in the global frame as well as manual guidance for the robotic arm to reach individual points.\\

This method can be effectively replaced with an online calibration process, where the robot identifies a set of keypoints in the camera image, executes a set of predetermined motions, and finally identifies the same set of keypoints after each motion whilst recalculating the relative camera position using this information. Thus, the objective of this project we aim to implement an algorithm which achieves complete automatic camera calibration from a sequence of uncalibrated images.\\

\section{Related Work}
Calibration of extrinsic camera parameters is a common problem in SLAM and has been addressed by many researchers in the past.\cite{b3} presents a method for extrinsic calibration of systems with two sensors within automotive setups. It provides an estimate for the transformation between two sensors, which is closely comparable to our problem statement.\\

Similarly, \cite{b2} is of particular interest as it presents an Extended Kalman Filter based approach for precisely determining the unknown transformation between a camera and an IMU. In our project however, we are concerned with the image sensor to end-effector transformation which is an analogous problem with a different application.\\

Regarding the estimation techniques used in this work, we adopt an Extended Kalman Filter approach, similar to that used in \cite{sfm}. This type of filter has already been extensively used in SLAM, and is particularly useful as the uncertainty in the estimated parameters is provided at every time step by computing their covariances, and can be used to represent the system fairly accurately as it moves around.\\

Since the environmental data acquired is neither streamlined nor is it immune to perspective changes it is imperative that unambiguous pattern instruments be deployed as calibration targets.\cite{b4} presented a qualitative performance evaluation of a few planar visual markers suitable for real-time applications. However, for the scope of our project we implement a QR code marker to efficiently pass the landmark identifier and camera to target depth data.\\

In traditional calibration approaches \cite{b8}, when the robot arm moves from the an existing pose to the new pose, a transform of both the arm positions relative to the base and the camera’s position relative to the target can be obtained. \cite{bfin} forms an important part of the recent stream of research on approaching eye-in-hand calibration techniques as several concepts are discussed in detail and  provides a comprehensive representation of the current state of art relevant to our project. Instead of using this traditional approach, our method directly uses Extended Kalman Filter to localize the pose of the camera, and thus calibrate the robotic arm directly.\\

Later on we explore alternative approaches as mentioned in \cite{sog}, \cite{ukf} to accurately model systems that involve a high degree of uncertainty and non-linearity such as with visual landmarks and dynamic environments.

\begin{figure}[!h]
\centering{\includegraphics[width=0.45\textwidth]{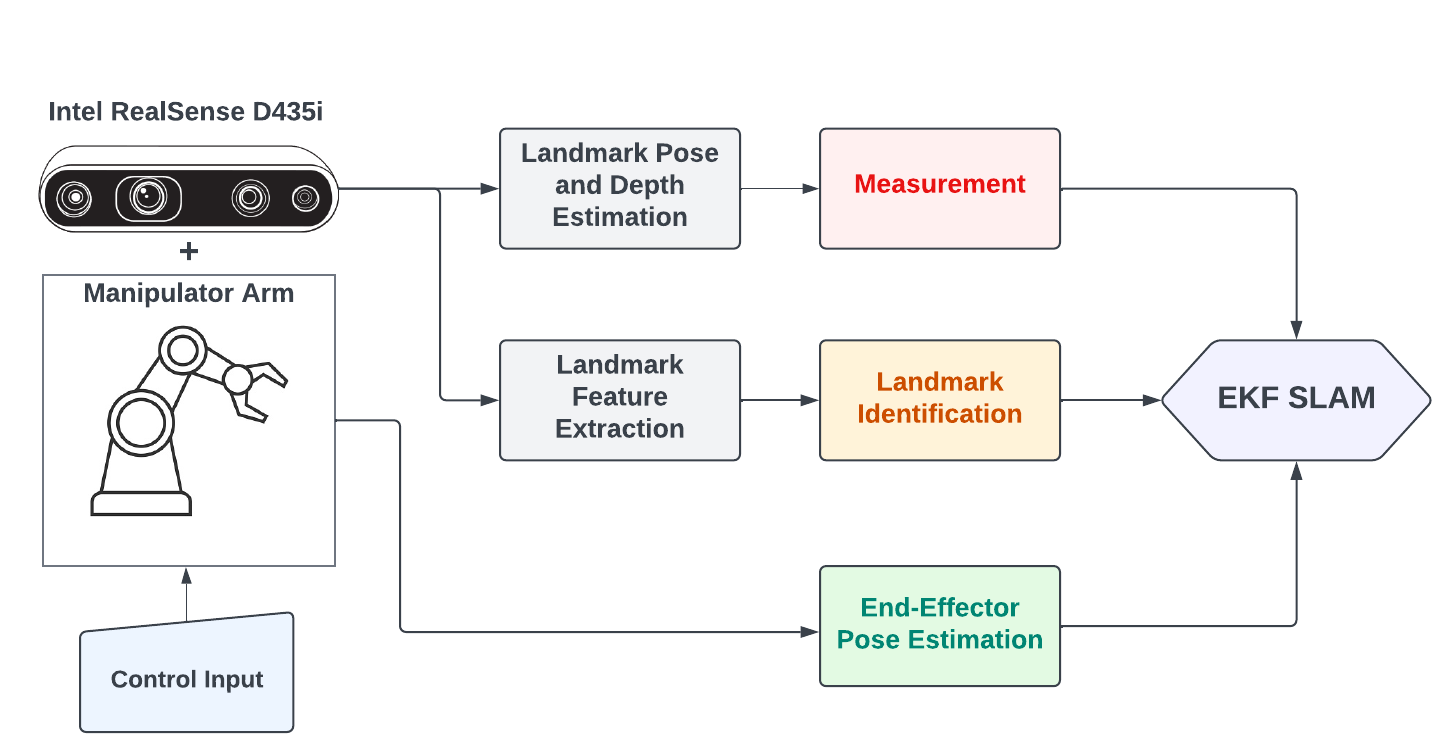}}
\caption{Eye-in-Hand calibration data flow}
\label{fig:chart}
\end{figure}

\section{Our Approach}

\begin{figure}[!h]
\centering{\includegraphics[width=0.48\textwidth]{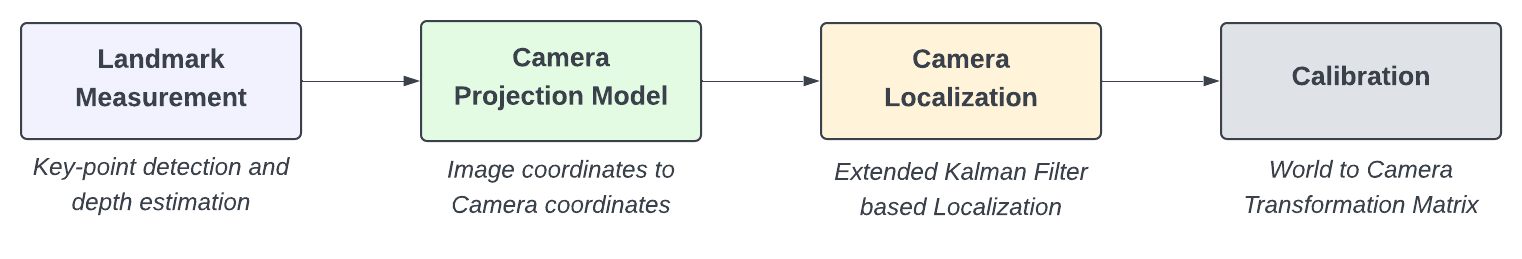}}
\caption{Different steps in Eye-in-Hand calibration using EKF}
\label{fig:approach}
\end{figure}

Our approach for eye-in-hand calibration using SLAM, can be broadly broken down into the steps shown in Fig. \ref{fig:approach}. We use an Extended Kalman Filter (EKF) based approach to localize the camera on the end-effector, which is predicted based on the control inputs to the end-effector and updated with our measurement data. For simplification, here the control inputs to the end-effector are omitted and we perform the localization with a pre-defined trajectory. The trajectory is defined by moving the end-effector to different locations in the 3D space, ensuring ample variations (atleast 20-40 mm) in the $x$, $y$ and $z$ coordinates of the base frame and all markers are visible in the field of view.. We employ feature markers for gathering the measurements using key point detection. The primary purpose of these markers are to pass relevant information such as depth data between the camera and target coordinate frames and additionally to allow easy landmark association. Use of such feature markers is a common practice in robotic-arm calibration.\\

Using a camera projection model the feature markers are scanned online, and the center of the feature markers are mapped to the camera coordinate system. These projected coordinates are subsequently used as landmark poses in the state vector of our localization algorithm, while the depth information from our sensor will provide the respective measurement data.

Finally, EKF based camera localization provides our estimate of the camera pose within the global frame which is used to infer the initial transformation matrix between the camera and end-effector positions. This matrix is iteratively updated as we arrive at new waypoints in the trajectory and as the sensor revisits previously mapped landmarks, the accuracy of our model incrementally improves.\\

\subsection{Landmark Measurement Module}
For the sensor measurement, we use the Intel Real sense D435i as our sensor for landmark position estimation, as this is one of the widely used sensors in robotics especially for pick and place applications. This specific sensor has an RGB camera and a depth camera augmented together which return depth data at an accuracy of 2\% up till 2 meters. We used a calibration template with a grid of multiple QR codes to aid our calibration. Using the calibration board also simplifies landmark identification, which helps with data association. Moreover, the QR codes allows us to make use of standard python libraries like Pyzbar \footnote{https://github.com/NaturalHistoryMuseum/pyzbar/} for identifying QR codes and we can embed the landmark ID within the QR code to overcome the data association problem. \\

For retrieving the landmark measurements in the camera frame we map the coordinates in the image plane $(u, v)$ to coordinates in camera frame of reference $(x_c, y_c, z_c)$. We do this by using the camera intrinsics and building the homography matrix in Eq. \ref{eq:homo} with it as follows. Therefore for the pre-processing, we essentially use the depth $z_c$ given by the sensor and the camera intrinsics to get the coordinates of the landmark in camera frame of reference which will be the output of our landmark measurement module.

\begin{equation}
    x_c=\frac{z_c (u-o_x)}{f m_x}
\end{equation}

\begin{equation}
    y_c=\frac{z_c (v-o_y)}{f m_y}
\end{equation}

rearranging the equations we get,\\
\begin{equation}
    x_c=\frac{z_c}{f m_x}u-o_x \frac{z_c}{f m_x}\\
\end{equation}

\begin{equation}
    y_c=\frac{z_c}{f m_y}v-o_y \frac{z_c}{f m_y}\\
\end{equation}

\begin{equation}
    \label{eq:homo}
    \begin{bmatrix}
        \frac{z_c}{f m_x}&0& -\frac{z_c}{f m_x} o_x\\
        0&\frac{z_c}{f m_y}& -\frac{z_c}{f m_y} o_y\\
        0&0&z_c
    \end{bmatrix}
    \begin{bmatrix}
    u\\
    v\\
    1\\
    \end{bmatrix}
    =
    \begin{bmatrix}
    x_c\\
    y_c\\
    z_c
    \end{bmatrix}
\end{equation}

where,
\begin{table}[h!]
\centering
\resizebox{0.45\textwidth}{!}{%
\begin{tabular}{ll} 
$(u, v)$ &= Landmark Coordinates in Image Plane \\
$(x_c, y_c, z_c)$ &= Landmark Coordinates in Camera Frame \\ 
$(O_x, O_y)$ &= Coordinates of image center \\ 
$f$ &= Camera Focal Length \\ 
$m_x$ &= Pixel density in x-direction \\
$m_y$ &= Pixel density in y-direction \\
\end{tabular}
}%
\end{table}

\begin{figure}[!h]
    \centering
    \includegraphics[width=0.3\textwidth]{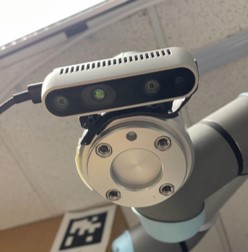}
    \caption{Intel Real sense D435i mounted on the end-effector for taking measurement}
    \label{fig:realsense}
\end{figure}
\begin{figure}[!h]
    \centering
    \includegraphics[width=0.45\textwidth]{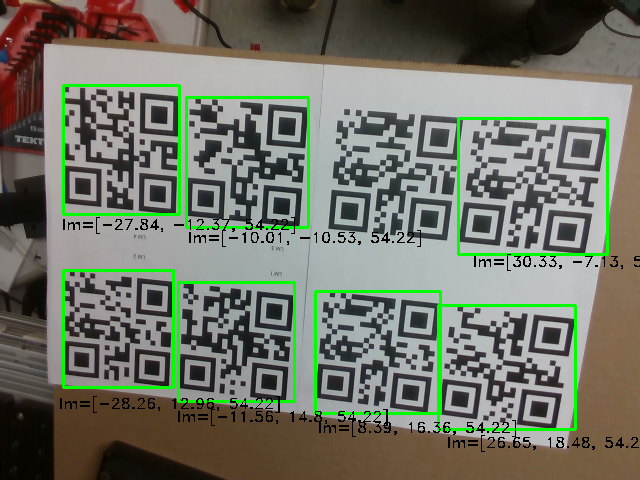}
    \caption{Results of data association and distance measurement (measurement values in centimeters)}
    \label{fig:measurement}
\end{figure}

Fig. \ref{fig:measurement} and Fig. \ref{fig:realsense} show the results from our landmark measurement module. Fig. \ref{fig:realsense} shows the camera observing the landmark (QR code) and in Fig. \ref{fig:measurement} we see the coordinates in camera frame of reference displayed which is the output of our measurement module. We will further use this to calculate bearing and range which will then be added to the EKF state vector.\\

\subsection{Camera Localization}
The most important component of our system is the camera localization. Among the different available methods (both online and offline) for localization such as filtering, pose-graph optimization, etc, we aim to set a baseline for the proposed method using Extended Kalman Filter (EKF) with known correspondences \cite{thrun2005probabilistic}. Because, despite the limitations of EKF it bodes well in this situation given the limited map size (feature marker calibration board), not a highly non-linear system and no computational constraints. Moreover, with EKF using feature-based map i.e. point landmarks, would suitably work with the center point measurements of the landmark markers.\\

The general Extended Kalman Filter paradigm can be summarised as prediction and update steps:
\\

\textbf{Prediction Step}
\begin{equation}
\centering
be\overline{li}ef(x_t) = \int p(x_t \vert x_{t-1}, u_t)     belief(x_{t-1}) dx_{t-1}
\end{equation}
\\

\textbf{Update Step}
\begin{equation}
\centering
belief(x_t) = p(z_t \vert x_t)  be\overline{li}ef(x_t) dx_{t-1}
\end{equation}

For our approach, assuming known landmark correspondences, the state $x_t$, representing the robot pose and the location of $k$ landmarks in 3D space, is\\
\begin{equation}
\centering
x_t = (p_t, m_{1,x}, m_{1,y}, m_{1,z} \dots, m_{k, x}, m_{k, y}, m_{k,z})
\end{equation}\\
where {$p_{t}$} is the pose of the end-effector comprising of its position and orientation in the 3D space in base frame: \\
\begin{equation}
    \centering
    p_t = (x_{t}, y_{t}, z_{t}, \phi_t, \theta_t, \psi_t)
\end{equation}
\\
Assuming the robot pose and the landmark locations are both Gaussian distribution, the probabilistic representation of the state is the mean and the variance:\\
\begin{equation}
\centering
\mu = (p_t, m_{1,x}, m_{1,y}, m_{1,z} \dots, m_{k, x}, m_{k, y}, m_{k,z})
\end{equation}

\begin{equation}
\Sigma = 
\begin{pmatrix}
\Sigma_{xx} & \Sigma_{xm} \\
\Sigma_{mx} & \Sigma_{mm} \\
\end{pmatrix}\\
\end{equation}\\

\textbf{Prediction Step:}\\

Here, we predict the expected robot state given the previous robot pose and the robot odometry.

We start by initialising $F_x$ to map from 6 dimension to (6 + 3$k$) dimension because the motion model $g$ only affects the robot motion and not the landmarks. The map is then used to predict expected state with our motion model and control inputs.

\begin{equation}
F_x = 
\begin{pmatrix}
I_{6\times6} & 0_{ 6\times3k}
\end{pmatrix} 
\end{equation}\

here the Jacobian for motion model is simply given by the equation below, where the Jacobian of our motion model $G_t^x$ is a simple identity matrix $I_3$.

\begin{equation}
G_t = 
\begin{pmatrix}
G_t^x & 0 \\ 
0 & I 
\end{pmatrix}
\end{equation}

Hereafter, we predict the covariance with Jacobian and motion noise.

\begin{equation}
\centering
\overline\Sigma = 
 G \Sigma G^T + F^T \Sigma_{control} F
\end{equation}\\

\textbf{Update Step:}\\

Here, we use the known data association to compute the Jacobian of H and proceed with computing the Kalman gain. Measurement Jacobians for pose and landmark: \\

\begin{equation} 
    \begin{split}
        H_{p} &= \begin{bmatrix}
            1 & 0 & 0 & H_{p00} & H_{p01} & H_{p02} \\ \\
            0 & 1 & 0 & H_{p10} & H_{p11} & H_{p12} \\ \\
            0 & 0 & 1 & H_{p20} & H_{p21} & H_{p22} \\
        \end{bmatrix}
    \end{split}
\end{equation}

\begin{equation} 
    \begin{split}
        H_{l} &= \begin{bmatrix}
            -1 & 0 & 0\\ \\
            0 & 0 & -1\\ \\
            0 & 1 & 0\\ 
        \end{bmatrix}
    \end{split}
\end{equation}

\

where,
\begin{itemize}[]
    \item[]$H_{p00}$ = {$s_{\phi}c_{\theta}x + (s_{\phi}s_{\theta}s_{\psi} + c_{\psi}c_{\phi})y + (s_{\phi}c_{\psi}s_{\theta} - c_{\phi}s_{\psi})z $}
    \item[]$H_{p01}$ = {$c_{\phi}s_{\theta}x - c_{\phi}c_{\theta}s_{\psi}y - c_{\phi}c_{\psi}c_{\theta}z$}
    \item[]$H_{p02}$ = {-$c_{\theta}$$y$ - $c_{\theta}$$z$}
    \item[]$H_{p10}$ = {$0$}
    \item[]$H_{p11}$ = {$c_{\theta}x + s_{\theta}s_{\psi}y + s_{\theta}c_{\psi}z$}
    \item[]$H_{p12}$ = {$-c_{\theta}c_{\psi}y + c_{\theta}s_{\psi}z$} 
    \item[]$H_{p20}$ = {$c_{\theta}c_{\phi}x - (s_{\phi}c_{\psi}-c_{\phi}s_{\theta}s_{\psi})y + (c_{\psi}c_{\phi}s_{\theta}+s_{\phi}s_{\psi})z$}
    \item[]$H_{p21}$ = {$-s_{\theta}s_{\phi}x + s_{\phi}c_{\theta}s_{\psi}y + c_{\psi}s_{\phi}c_{\theta}z$}
    \item[]$H_{p22}$ = {$(-c_{\phi}s_{\psi} + s_{\phi}s_{\theta}c_{\psi})y - (s_{\psi}s_{\phi}s_{\theta} + c_{\phi}c_{\psi})z$}
\end{itemize}\

\

where, $s$ and $c$ are the $sine$ and $cosine$ functions, and $x, y, z$ are the control inputs for the end-effector positions.

\

By combining the computed Jacobians above, we arrive at the overall update Jacobian which is used to calculate the Kalman gain

\begin{equation}
    K_{t} = \Sigma_{t}H_{t}^T (H_{t}\Sigma_{t}H_{t}^T + Q_{t})^{-1}
\end{equation}

where, 
\begin{equation}
H_{t} = \begin{bmatrix}
H_{p} & H_{l}
\end{bmatrix}  \quad , \quad
Q_{t} = \begin{bmatrix}
\sigma_{x}^2 & 0 & 0\\ \\
0 & \sigma_{y}^2 & 0\\ \\
0 & 0 & \sigma_{\alpha}^2
\end{bmatrix}
\end{equation}
\\

The computed Kalman gain is then weighted against the predicted state and covariance matrices to correctly update the localization.\\

Although the uncertainties in our approach will be modeled using normal distribution $\mathcal{N} (\mu,\,\Sigma)$ , but they could also be modeled through Sum of Gaussians (SoG), as implemented in \cite{sfm} to account for any large non-linearities which may be introduced in completely arbitrary placement of camera on the end-effector. Although we are not employing SoG \cite{sog}, in the SoG approach, the probability density function of the estimated parameters $p(x)$ could be then approximated by a weighted sum of multivariate Gaussians as:
\begin{equation}
    p(x) = \sum_{{i=1}}^{n_g} \alpha^{(i)} \mathcal{N} (\mu,\,\Sigma)\,
\end{equation}
where $n_g$ stands for the number of Gaussians, $\mu$ and $\Sigma$ represents the mean and covariance matrix for each Gaussian and $\alpha^{(i)}$ its corresponding positive weight.\\

The implementation of EKF in our approach is analogous to a robot localization problem in 3D space, where the Robot arm and RGB-D Camera need to be localized with respect to the QR markers. An underlying assumption here is that the robot arm is fully calibrated, so the end-effector position would be fairly accurate. Additionally, at each way point in the pre-defined trajectory as mentioned earlier, we ensure that all the landmarks are within the camera's field of view.\\

\subsection{Calibration}
As stated earlier, we make a reasonable assumption that the robot arm is fully calibrated with respect to the end-effector position due to industry standards of robot arms. We assume that the transform from the camera’s coordinates to the end-effector’s coordinates stays the same, as the camera is located at a fixed position on the end-effector. We obtain an equation based on the poses of the vectors and solving this equation, we can obtain the transform between the camera and the arm being measured, which is the final calibration that we are looking for. Although, our method doesn't explicitly solve the equations \ref{eq:c1} to \ref{eq:c2}, the equations are representational to the transform being calculated.
\begin{figure}[!h]
    \centering
    \includegraphics[width=0.4\textwidth]{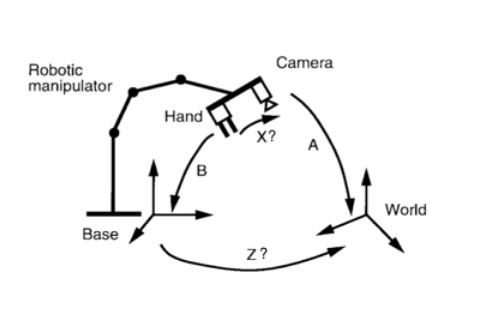}
    \caption{Traditional approach in finding transform between the camera and end-effector}
    \label{fig:calib}
\end{figure}

The mathematical formulation of the calibration problem shown below translates directly to the batch processing algorithm we have implemented. However it is also an implicit representation of the transform being calculated using EKF.
\begin{equation}
\label{eq:c1}
    AX = XB
\end{equation}
where,
\begin{equation}
    A = A_{i+1}A_i^{-1}
\end{equation}
\begin{equation}
\label{eq:c2}
    B = B_{i+1}B_i^{-1}
\end{equation}
in which,
\begin{itemize}[]
    \item[]$A_i$ = Previous camera-to-landmark transformation
    \item[]$A_{i+1}$ = New camera-to-landmark transformation
    \item[]$B_i$ = Previous world origin-to-end-effector transform
    \item[]$B_{i+1}$ = New world origin-to-end-effector transform
    \item[]$X$ = Transform between the camera and arm
\end{itemize}
\
\\
In our approach, we obtain the transform with camera localization instead. This allows us to directly proceed to perform calibration of the camera's pose with respect to the world's frame with the transformation:

\begin{equation}
    X_c = M_{ext}X_w
\end{equation}
where, \\
\begin{itemize}[]
    \item[]$X_c$ = Estimation of camera's pose relative to end-effector
    \item[]$M_{ext}$ = Extrinsic matrix of camera's transformation from world frame
    \item[]$X_w$ = Calibrated camera's position in world frame
\end{itemize}

\section{Experimental Setup}
Our hardware test setup includes three components; an in-house Universal Robots  (UR5) robotic manipulator arm with 6 degrees of freedom, an Intel real-sense D-435i depth camera, and a set of eight Quick Response (QR) markers which behave as static landmarks. 

\begin{figure}[!h]
    \centering
    \includegraphics[width=0.4\textwidth]{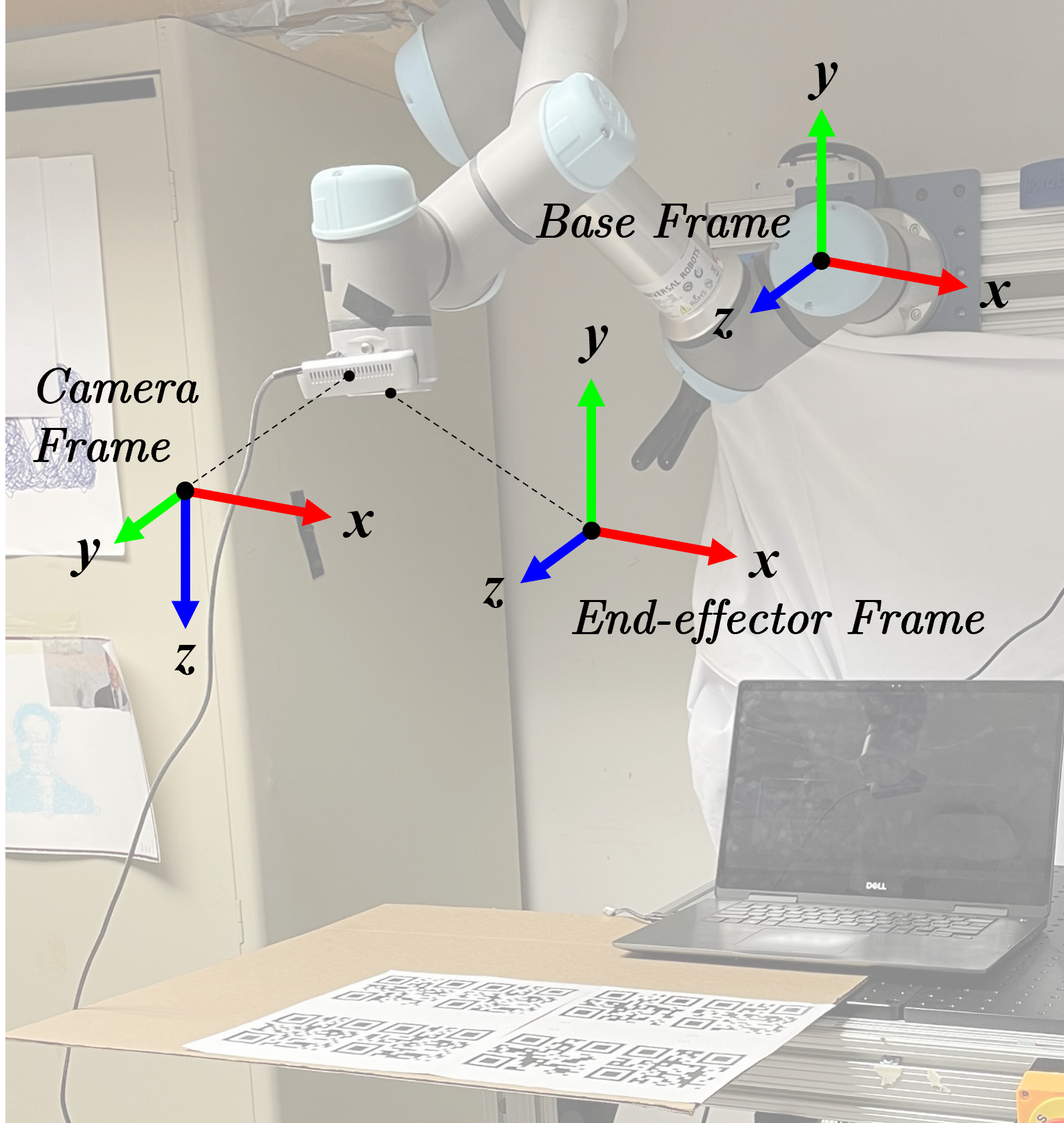}
    \caption{Coordinate Frames in the experimental setup}
    \label{fig:frames}
\end{figure}

\begin{figure}[!h]
    \centering
    \includegraphics[width=0.4\textwidth]{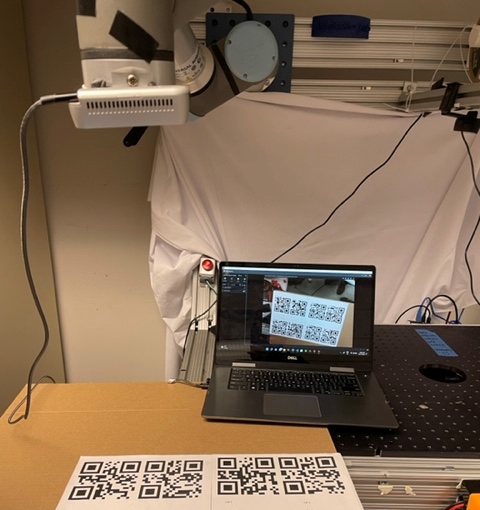}
    \caption{Experimental Setup showing the QR Marker calibration board (bottom) and robot arm with D435i mounted on it (top).}
    \label{fig:setup}
\end{figure}

As seen in Fig. \ref{fig:setup}, the UR5 manipulator is mounted onto the wall, while the camera is seen mounted perpendicularly on the end-effector. We define the positive Z-axis within the base frame and the end-effector frame as projecting from the wall towards the user, the Y-axis as a vertical-upward traversal (downward for negative Y), and the X-axis as horizontal-right traversal (left for negative X). The camera frame is oriented differently from the base frame, as can be seen in Fig. \ref{fig:frames}. \\

The camera is placed at a predetermined location on the robot end-effector approximately 6mm along the positive Y-axis and 40 mm along the positive Z-axis from the end-effector frame. The camera provides the image data for the key-point detection and feature extraction to run the localization and calibration algorithms, while the QR markers are placed in a planar fashion within the X-Z plane of base frame as shown in Fig. 4.\\

The dataset used to test our algorithm is generated by traversing through a set of predetermined way-points within the 3D robot workspace. These way-points are provided via a teach pendant and indicates the position of the end-effector tool-centre with respect to the robot base-frame.\\

It is imperative that we understand a few assumptions intrinsic to our experimental setup in order to test our algorithm and validate our results. Firstly, we assume that the UR5 manipulator arm is fully calibrated and the joint motor-control inputs have negligible errors. Secondly, we assume that the camera-end-effector transformation "X" remains the same throughout the robot motion. These are reasonable assumptions considering that in an industrial setting, these parameters are largely invariant.\\

With regard to our measurements, we have approached our experiment considering that the camera captures all eight QR markers at every way-point.\\

\begin{figure}[!h]
    \centering
    \includegraphics[width=0.45\textwidth]{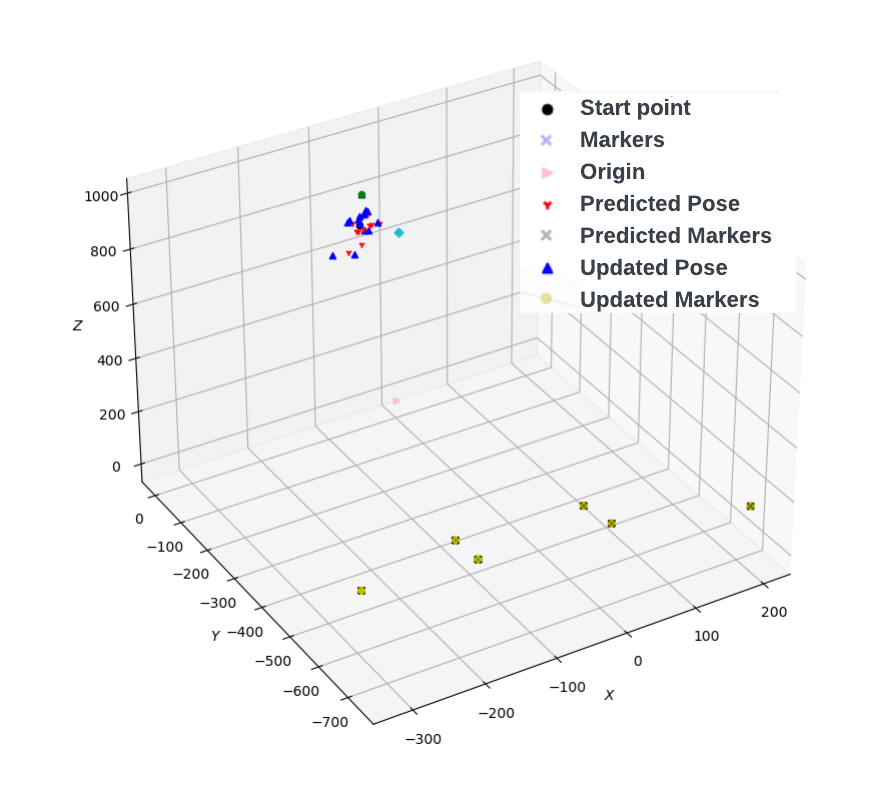}
    \caption{EKF Localization in 3D space}
    \label{fig:sim}
\end{figure}

\section{Results}
Using the data collected from the experimental setup, we used our EKF SLAM algorithm to localize the camera and then perform calibration. From the EKF algorithm, we finally get the location of the camera in the base frame, is assumed to be at the base of the robot. The localization results can be seen in table \ref{tab:results}. The camera was mounted at an off-set location i.e. not exactly at the end-effector center and also due to the camera center being inside the camera. This offset could only be calculated approximately, due to the geometry of the end-effector which actually is a clear indication of why calibration is necessary. The approximate off-set between the end-effector center and the camera center can be seen in table \ref{tab:results}. When this offset is taken into account with the final location of the end-effector, it approximately gives us the ground truth location of the camera. Consequently, the comparison of our localization and this approximate ground truth indicates that our localization was quite accurate in the $x$ and $z$ directions, with an absolute error of 0.7\% and 2.5\% respectively. Along, the y direction we got incorrect localization with a 46.39\% absolute error. There could be many reasons behind this, we believe the high error could be attributed to the increased error in D545i's depth estimation as the distance from the camera increases. Also, it is likely that during the data collection, the camera could have moved owing to influence of gravity and some inadvertant tugs to the cable connecting to D435i for data transfer.

\begin{table}[h!]
\centering
\caption{Localization Results}
\resizebox{0.45\textwidth}{!}{%
 \begin{tabular}{|l|c|c|c|} 
 \hline
Coordinate &$x$ &$y$ &$z$ \\ \hline
End-effector Actual (mm) &-102.12 &-265.85 &955.47 \\ 
Camera Off-set (mm) &0 &6 &40 \\ 
Camera Actual (mm) &-102.12 &-259.65 &995.47 \\ 
Camera Localization (mm) &-102.83 &-142.52 &980.34 \\ \hline
Absolute Error &0.70\% &46.39\% &2.60 \\ \hline
L2 Distance Error (mm) &\multicolumn{3}{|c|}{125.82} \\ \hline
\end{tabular}
}%
\label{tab:results}
\end{table}

As a result of both EKF and Batch processing implementation we get a localized state vector with rotation vector and X,Y, and Z co-ordinates stacked as the state vector. The rotation vector is converted into a 3x3 Rotation matrix using the Rodrigues formula and is stacked with the translation vector to get a homogenous transformation as seen in equation \ref{eq:11}, the resulting matrix directly maps coordinates in world frame to camera frame and is the result of our calibration.\\
\begin{equation}
\label{eq:11}
 \begin{bmatrix}
        \alpha\\
        \beta\\
        \gamma\\
        t_x\\
        t_y\\
        t_z\\
    \end{bmatrix}
    \Longrightarrow
    \begin{bmatrix}
        R&t\\
        0&1\\
    \end{bmatrix}
\end{equation}\\
\begin{equation}
    \begin{bmatrix}
        X_c\\
        Y_c\\
        Z_c\\
        1\\
    \end{bmatrix}
    =
    \begin{bmatrix}
    r_{11}&r_{12}&r_{13}&t_x\\
    r_{21}&r_{22}&r_{23}&t_y\\
    r_{31}&r_{32}&r_{33}&t_z\\
    0&0&0&1\\
    \end{bmatrix}
    \begin{bmatrix}
    X_w\\
    Y_w\\
    Z_w\\
    1\\
    \end{bmatrix}
\end{equation}\\
\section{Conclusion and Future Work}
From our experiment, we learn that EKF based approaches are well suited to solve the online robotic arm calibration problem. Although EKF approaches may time some time to process when given large number of landmarks, it is appropriate in applying to this problem, as the robotic arm calibration process doesn't necessarily require online localization, and the data can also be processed offline, which was demonstrated through results of our approach. As shown in our results, the EKF approach worked well in our experiment for x and z directions, despite the crude experimental setup, albeit there was a significant error in the y direction. Our approach when implemented with automatic centering of robot arm, could be used without any human intervention thus making calibration a fast and robust process. This would make our approach a fully-automatic calibration process.\\

Considering future work related to eye-in-hand calibration of real industrial settings, visual landmarks of the environment could be used instead of QR codes as landmarks, which would allow for the robot arm to be calibrated without a calibration board. As visual landmarks would introduce much more non-linearity in the process compared to our calibration board of QR codes, EKF might underestimate the covariance of the function. Therefore, alternative approaches, such as Iterated Extended Kalman Filter (IEKF), Unscented Kalman Filter (UKF) \cite{ukf}, could be used to deal with higher non-linearities and the Sum of Gaussians (SoG) approach \cite{sog} can be used to model uncertainty for accurate calibration. Another alternative approach that we have worked on as a stretch goal, nevertheless without presentable results, is the factor graph based optimization. The factor graph based approach is suitable in this application owing to there being no need of online localization and also it would make the entire calibration algorithm more straightforward.

\medskip
{\small
\bibliographystyle{IEEEtran}
\bibliography{references}
}

\end{document}